# A Latent Risk–Aware Machine Learning Approach for Predicting Operational Success in Clinical Trials based on TrialsBank™

Iness Halimi[1,†], Emmanuel Piffo[1,†,*], Oumnia Boudersa[2], Yvan Marcel Carre Vilmorin[3], Melissa Ait-ikhlef[4], Karima Kone[4], Andy Tan[4], Augustin Medina[4], Juliette Hernando[1,4,5], Sheila Ernest[6], Vatche Bartekian[6], Karine Lalonde[1,7], Mireille E Schnitzer[4,8,9], Gianolli Dorcelus[1,3].

[1] Groupe Sorintellis Inc., Montréal, Québec, Canada
[2] Department of Computer Science and Operations Research, Université de Montréal, Montréal, Québec, Canada
[3] Department of Computer Engineering, École Polytechnique de Montréal, Montréal, Québec, Canada
[4] Faculty of Pharmacy, Université de Montréal, Montréal, Québec, Canada
[5] Faculty of Pharmacy, Université Laval, Québec City, Québec, Canada
[6] Vantage BioTrials, Montréal, Québec, Canada
[7] CDS monitoring, Montréal, Québec, Canada
[8] Department of Social and Preventive Medicine, School of Public Health, Université de Montréal, Montréal, Québec, Canada
[9] Department of Epidemiology, Biostatistics and Occupational Health, McGill University, Montréal, Québec, Canada
† These authors contributed equally to this work.
* Corresponding author: Emmanuel Piffo (ep@sorintellis.ai)

## Abstract

Clinical trials are characterized by high costs, extended timelines, and substantial operational risk, yet reliable prospective methods for predicting trial success before initiation remain limited. Existing artificial intelligence approaches often focus on isolated metrics or specific development stages and frequently rely on variables unavailable at the trial design phase, limiting real-world applicability. We present a hierarchical latent risk-aware machine learning framework for prospective prediction of clinical trial operational success using a curated subset of TrialsBank, a proprietary AI-ready database developed by Sorintellis, comprising 13,700 trials. Operational success was defined as the ability to initiate, conduct, and complete a clinical trial according to planned timelines, recruitment targets, and protocol specifications through database lock. This approach decomposes operational success prediction into two modeling stages. First, intermediate latent operational risk factors are predicted using more than 180 drug- and trial-level features available before trial initiation. These predicted latent risks are then integrated into a downstream model to estimate the probability of operational success. A staged data-splitting strategy was employed to prevent information leakage, and models were benchmarked using XGBoost, CatBoost, and Explainable Boosting Machines. Across Phase I-III, the framework achieves strong out-of-sample performance, with F1-scores of 0.93, 0.92, and 0.91, respectively. Incorporating latent risk drivers improves discrimination of operational failures, and performance remains robust under independent inference evaluation. These results demonstrate that clinical trial operational success can be prospectively forecasted using a latent risk-aware AI framework, enabling early risk assessment and supporting data-driven clinical development decision-making.

# Introduction

Clinical drug development process is costly, time-intensive, and risky. The global clinical trials market was valued at approximately USD 90.1 billion in 2025 and is projected to reach around USD 123.5 billion by 2030, representing a compound annual growth rate (CAGR) of about 6.5%[1]. In parallel, clinical drug development timelines from Phase I initiation to regulatory approval remain substantial, typically spanning close to a decade, with recent analyses estimating an average duration of approximately 9 years[2,3]. Despite escalating costs and lengthening development timelines, approximately 90% of drug candidates entering Phase I ultimately fail to reach the market[4]. Moreover, only a minority of approved products generate positive return on investment[5,6,7]. Collectively, these challenges highlight a critical gap in clinical development practice: the lack of robust, data-driven frameworks capable of predicting clinical trial success a priori in the development lifecycle. As highlighted by industry experts, clinical development programs are governed through structured phase-specific decision gates, where advancement decisions integrate scientific evidence, operational feasibility, safety profile evolution, competitive landscape, and capital allocation constraints. As such, risk and success are inherently context-dependent, varying by therapeutic area, modality, population complexity, geographic footprint, and execution model. The ability to anticipate execution risk early (i.e. before first patient in) has therefore become increasingly central to program governance, portfolio prioritization, and risk-based oversight planning.

Against this backdrop, artificial intelligence (AI) is increasingly recognized as a key enabler for improving decision-making across the clinical drug development lifecycle. Reflecting this momentum, the global AI-in-clinical-trials market is projected to grow from approximately USD 1.35 billion in 2024 to USD 2.74 billion by 2030, corresponding to a CAGR of 12.4%[8]. This rapid expansion underscores the growing reliance on data-driven approaches to not only optimize trial design and operational efficiency, but also to enable earlier feasibility assessment, proportionate oversight planning and structure risk mitigation consistent with Risk-Based Quality Management (RBQM)[9,10] expectations.

**Emergence of artificial intelligence in clinical drug development workflows**

On the one hand, artificial intelligence has been increasingly integrated into the scientific and operational workflows of clinical drug development, where it is primarily applied to optimize discrete components of the process. Recent advances[11,12] have demonstrated that artificial intelligence can be directly embedded within digital twin architectures to generate patient-level virtual representations capable of forecasting clinical trajectories. In particular, AI-generated digital twins have been proposed as a scalable and regulatory-aligned approach to optimize clinical trial design, reduce sample size, and improve statistical efficiency[13]. In parallel, deep learning has been developed to predict clinical trial outcomes, defined as whether a trial met its primary endpoint(s), by integrating HINT multimodal data related to the investigational compound, trial eligibility criteria, and disease characteristics[14]. Artificial intelligence has also been applied to predict adverse events from sparse clinical trial data, enabling early identification of safety risks that may only emerge post-approval. Graph-based Machine Learning (ML) approaches have demonstrated the ability to infer drug–adverse event associations and support proactive safety assessment during clinical development[15]. In addition, AI-driven methods have been proposed to optimize patient recruitment by learning from heterogeneous investigator- and trial-level data to rank investigators according to expected enrollment performance, demonstrating strong potential to reduce recruitment delays and associated operational costs[16]. Indeed, a recently developed machine learning algorithm forecasts five efficiency metrics independently including screen failure ratio, dropout ratio, pre-enrollment duration, enrollment duration and study duration using a total of 23 trial design features characterizing 2050 clinical trials conducting by pharmaceutical company Roche[17].

**Artificial intelligence for predicting clinical trial outcomes**

On the other hand, a growing body of work has leveraged artificial intelligence to predict downstream clinical trial outcomes, thereby providing a more holistic assessment of the clinical development process. Lo et al. 2018[18] applied machine learning models to commercial *Informa* dataset to predict regulatory approval from Phase II and Phase III trials using approximately 30 predictive features. Vergetis et al. 2021[19] introduced an oncology-focused Probability of Technical and Regulatory Success (PTRS) model based on around 117 variables derived from a proprietary dataset, while Feijoo et al. 2020[20] leveraged public registries (*ClinicalTrials.gov* via AACT) and commercial *BioMedTracker* database to develop machine learning models to predict phase transitions from Phase II to Phase III and from Phase III to approval. In addition, Aliper et al. 2023[21] introduced *inClinico*, a transformer-based AI framework designed to predict phase transitions of drug–indication pairs from Phase II to Phase III. The model leverages proprietary multimodal datasets curated by human experts and augmented using a generative large language model GPT-3.5. More recently, Reinisch et al. 2024[22] proposed a large language model–based framework to predict clinical trial phase transitions across all phases through approval, using integrated proprietary datasets combining *ClinicalTrials.gov* and *BioMedTracker*. Beyond phase transition and regulatory outcome prediction, Ferdowsi et al. 2023[23] investigated the use of deep learning to estimate clinical trial risk based on protocol-level characteristics. By leveraging features such as protocol amendments, study status, enrollment drop, and trial duration, the authors proposed a retrospective risk labeling strategy to categorize trials into low-, medium-, and high-risk groups[23]. Their approach relies on an ensemble architecture combining transformer models with graph neural networks, enabling the capture of complex dependencies between protocol attributes and associated risk profiles[23].

**Limitations of existing AI-driven approaches for clinical trial success prediction**

Despite significant progress, existing AI-driven approaches for clinical trial predictive modeling exhibit several fundamental limitations that constrain their ability to support holistic, early-stage decision-making in real-world clinical development.

- Existing studies on artificial intelligence for predicting clinical trial outcomes often lack a pragmatic conceptualization of clinical trial success. Indeed, from a practical perspective, and depending on the stakeholders involved including sponsors, contract research organizations (CROs), investigational sites, regulatory agencies and participants, the notion of success is inherently multidimensional and context dependent. Accordingly, clinical development is governed by multiple stage-specific success indicators that are sequential, interdependent, variably prioritized across stakeholders, and require continuous monitoring and adaptive decision-making throughout each stage of the development process. However, much of the literature[18–22] addressing the prediction of clinical trial outcomes reduces success to a single criterion, most commonly regulatory approval or transition to a subsequent clinical phase. Such simplification fails to reflect the operational realities of drug clinical development. Regulatory approval represents a terminal milestone occurring late in a long and complex development process and remains contingent upon the successful completion of numerous upstream stages. Reducing success to a single global endpoint therefore obscures this diversity and limits alignment with real-world decision-making needs.

  Clinical trial success relies fundamentally on operational success, which represents the proximal prerequisite for generating high-quality, interpretable evidence package within anticipated timelines and budget. Operational inefficiencies in clinical trials have significant financial and strategic implications for drug development programs. In this context, it is estimated approximately 85% of clinical trials experience delays, with 94% extending beyond one month, and that such delays may cost sponsors an estimated USD 600,000 to 8 million per day in lost value[24]. In real-world development programs, operational performance functions as a leading indicator of downstream program viability, even though scientific or strategic factors may independently determine final development outcomes. Accordingly, operational success should be understood not as a guarantee of regulatory approval, but as a necessary enabler of decision-useful clinical evidence. Additionally, from a sponsor and CROs perspectives, operational success is typically evaluated through measurable execution dimensions including site activation performance, recruitment and retention relative to assumptions, protocol adherence, safety management robustness, timeline variance, and readiness for database lock and regulatory inspection.

- In addition, prior works often overlook the complex causal relationships and interdependencies between variables when predicting clinical trial success. While such approaches may be of theoretical interested and useful for optimizing specific components of clinical trials, they are often insufficiently aligned with the operational realities and decision-making needs of the pharmaceutical industry. Indeed, a large proportion of existing methods focus on predicting or optimizing isolated components of the clinical development process, such as patient recruitment efficiency, adverse event occurrence, or protocol amendment frequency, without integrating these dimensions into a unified notion of overall trial success. By neglecting the interactions and causal dependencies among clinical, operational, and strategic variables, these models struggle to deliver truly actionable insights capable of informing key clinical development decisions, such as trial design optimization, portfolio prioritization, or early risk mitigation. For example, protocol complexity directly influences recruitment feasibility, participant burden, dropout risk, and protocol deviation rates. Similarly, site capability, competing trials, and country-level regulatory timelines affect enrollment performance, budget burn, and overall program duration. Safety profile uncertainty may drive conservative eligibility criteria, which in turn increases screen failure rates and recruitment pressure. These interdependencies underscore the importance of constructing sufficiently contextualized, rich, and finely annotated datasets curated by experts with deep practical knowledge of clinical trial design, planning, management, and operational execution.

- Third, and most critically from an operational standpoint, many existing models rely on historical, fully observed datasets in which key drivers of trial success are retrospectively available. However, in real-world clinical development, several high-impact operational drivers, such as recruitment rate, dropout ratio, or protocol deviations, cannot be observed or quantified prior to trial initiation. Rather than being treated as observable inputs or omitted altogether, these drivers should be explicitly modeled as latent variables. Models that fail to account for this constraint risk achieving strong retrospective performance while remaining non-operational for prospective trial design and early risk mitigation.

**Data limitations for AI-driven clinical trial modeling**

Beyond modeling limitations, the practical adoption of AI in clinical development is further constrained by the availability and the structure of ML-ready data. Reflecting this reality, Gartner estimates that up to 60% of AI initiatives built on non–AI-ready data will ultimately be abandoned by 2026[25]. While large volumes of clinical trial data exist across public registries and internal sponsor systems, these data are rarely curated in a manner that directly supports prospective, decision-oriented modeling. As emphasized in prior work, AI systems can only be as strong as the data that fuels them (*garbage in, garbage out*[26]), underscoring the central role of data quality in building trustworthy models[27]. Moreover, widely used public registries such as *ClinicalTrials.gov*, which are frequently leveraged as primary data sources for AI model development or as foundations for proprietary databases, present structural limitations that further complicate their use for predictive modeling:

- First, *ClinicalTrials.gov* includes multiple study types beyond interventional clinical trials, such as observational studies and expanded access records, each governed by different reporting requirements and data availability, leading to heterogeneous and non-comparable records[28]. In addition, industry experts report that data completeness and update frequency vary across sponsors and study types, resulting in incomplete outcome reporting. Registry entries may lack granular operational detail (e.g., site activation timelines, monitoring models, vendor configurations, protocol amendment history, or country-level execution dynamics) that are essential for prospective operational risk modeling.
- Second, because follow-on, extension, and related studies are often registered as separate and unlinked records, reconstructing full development trajectories or understanding cross-trial dependencies remain challenging[28]. This often requires non-trivial manual harmonization, which, if not carefully performed, may increase the risk of fragmentation and information leakage in predictive modeling workflows.
- Third, registry entries are subject to modifications over time, with evolving reporting standards and delayed updates, introducing temporal inconsistencies that complicate reproducibility and longitudinal analyses[28].
- Finally, many records contain missing, incomplete, or weakly standardized data elements, reflecting changes in mandatory reporting requirements and variable compliance, which collectively constrain the reliability and downstream usability of *ClinicalTrials.gov* data for AI-driven clinical trial success modeling[28].

Public registries remain essential transparency instruments; however, their structure and reporting design make them necessary but insufficient foundations for prospective, decision-oriented operational forecasting without additional curation, contextualization, and expert validation.

As describe above, despite growing interest in predictive modeling of clinical trial outcomes, prior research has largely depended on fragmented datasets characterized by inconsistent feature coverage across studies[18–22]. The absence of a public, exhaustive source enumerating the determinants of clinical trial success has limited cross-study comparability and generalizability. Indeed, multiple authors have underscored the significant challenges involved in constructing high-quality historical datasets for clinical trial success modeling. In particular, acquiring reliable and comprehensive data on clinical trial features and outcomes is often costly, time-consuming, and susceptible to inconsistencies, thereby limiting the robustness and scalability of predictive approaches, which frequently depend on manual curation from heterogeneous and partially structured data sources[29]. Beyond data availability, interpretation at scale represents an equally critical bottleneck. Clinical trial protocols encode scientific hypotheses, medical assumptions, operational constraints, and regulatory considerations within highly technical documents. As noted by Reinisch et al. 2024[22], accurate comprehension of these protocols often requires deep, domain-specific clinical expertise[22]. Despite growing efforts in the literature to leverage the potential of large language models for building clinical trial databases and for extracting and structuring information from complex trial documents, their use for fully automated annotation remains fundamentally constrained. In particular, prior studies have highlighted limitations related to hallucinated outputs, sensitivity to prompt formulation, limited mechanisms for knowledge updating, and insufficient transparency, which collectively necessitate expert validation and human-in-the-loop frameworks to ensure data reliability and regulatory compliance in real-world clinical development settings[30].

**Proposed framework**

To address these limitations, in this article we propose a data-centric, hierarchical latent risk-aware machine learning framework for the prospective prediction of clinical trial operational success. Our approach explicitly accounts for the multidimensional and interdependent nature of clinical trial risk, the existence of stage-specific success metrics, and the constraints on the availability of pre-trial feature values. Built on AI-ready data representations derived from TrialsBank™, the framework is designed to support forward-looking, operationally actionable decision-making by aligning predictive outputs with real-world clinical development practices.

## Materials and Methods

### Data Source: TrialsBank™ Clinical Trial Database

For the present study, the machine learning models were developed using a curated analytical subset of TrialsBank™, a proprietary, AI-ready dataset developed by Sorintellis. The dataset comprises 13,700 unique Phase I–III clinical trials, derived through a validated curation and sampling protocol involving expert quality verification to ensure data completeness and standardized variable availability. The dataset includes trials mapped to more than 2000 unique therapies and 21 therapeutic areas. TrialsBank™ was explicitly designed to address the data constraints that have historically limited predictive modelling in clinical trials through large-scale data structuring, expert annotation, multimodal integration, and longitudinal reconstruction of clinical drug development pathways[31]. In particular, TrialsBank™ captures a comprehensive, context-aware and multidimensional risk landscape integrating scientific, pharmacologic, biostatistical, epidemiologic, operational, regulatory, economic, and financial dimensions through a methodology that integrates scientific literature review with in-depth, semi-structured interviews across a broad spectrum of industry stakeholders. The literature review[32] focused on identifying determinants of clinical trial success and failure across all therapeutic areas over a 20-year period (2002–2022), applying a structured search strategy across MEDLINE, Embase, and Web of Science, and initially identifying 3,715 potentially relevant publications. Complementarily, 20 semi-structured interviews were conducted with pharmaceutical experts, including large pharmaceutical companies, biotechnology firms, and CROs, each possessing a minimum of 10 years of experience, using a validated interview guide comprising 15 predefined questions to both confirm factors identified in the literature and elicit additional determinants grounded in real-world clinical trial practice. This dual approach resulted in the definition, harmonization, and validation of a standardized and exhaustive taxonomy of risk factors and key success outcomes across the clinical development lifecycle, explicitly grounded in both scientific evidence and real-world operational practice. By integrating this standardized risk taxonomy into the dataset, TrialsBank™ enables predictive and causal modeling of clinical trial outcomes.

Given the critical role of human expertise in ensuring the validity and operational relevance of complex clinical trial annotations, TrialsBank™ was constructed using a human-expert-in-the-loop curation framework. Over a four-year period beginning in 2022, raw clinical trial data aggregated from multiple public and private data sources were systematically curated and annotated by a multidisciplinary team of more than 30 pharmaceutical domain experts. These experts represented complementary areas of expertise, including and not limited to clinical research development, clinical trial operations, biostatistics, clinical trial design, regulatory affairs, pharmacology, clinical trial management, health economics and outcomes research, pharmacovigilance, market access, and medical affairs.

To ensure the scientific validity, and clinical relevance of annotated drug trials, annotation procedures were governed by predefined and validated standardized operating procedures (SOPs) designed to ensure consistency, traceability, and reproducibility across annotators. These SOPs defined annotation scopes and specified, for each variable, its formal definition, annotation rules, and the authoritative data sources to be used. The construction of TrialsBank™ relied on the integration of approximately 15 structured and unstructured trustworthy data sources[33] (e.g., ClinicalTrials.gov, EudraCT, Health Canada's drug product database, Drugs@FDA, Pharmaceutical company pipelines, Clinical research organization databases), enabling harmonized interpretation of heterogeneous clinical trial information while preserving domain-specific contextualization. Following annotation, a rigorous quality control and validation process was conducted through cross-review by independent domain experts, ensuring accuracy, internal consistency, and alignment with clinical practice.

The resulting curated analytical subset used in the present study comprises 182 expert-curated features organized into seven major categories capturing the multidimensional determinants of clinical trial performance. The following list presents these categories, along with selected illustrative examples of features within each category:

- Investigational therapy characteristics (e.g., biological target family, mechanism of action, therapeutic modality).
- Disease and indication attributes (e.g., therapeutic area, rare disease status, indication prevalence).
- Trial design and protocol characteristics (e.g., randomization scheme, number of comparators, planned number of visits, planned enrollment size).
- Outcome and endpoint specifications (e.g., primary endpoint, secondary endpoints, statistical consideration).
- Study participant characteristics (e.g., inclusion criteria, exclusion criteria).
- Sponsor characteristics (e.g., sponsor type, sponsor experience in the indication).
- Site-level characteristics (e.g., participating countries, number of investigational sites, regional distribution).

Beyond individual trial records, TrialsBank™ links all clinical studies associated with a given investigational pharmaceutical product, reconstructing complete development trajectories from first-in-human studies through regulatory submission and market authorization. This unified and normalized framework enables cross-trial analysis at scale. The comprehensive feature set, expert-curated annotations, and longitudinal continuity of the database enable both descriptive analytics and advanced predictive modeling across the clinical drug development lifecycle.

The resulting high-quality dataset comprises curated, context-aware, AI-ready clinical trial records that supports longitudinal analysis for both retrospective reconstruction of clinical development trajectories and prospective forecasting of drug development program evolution.

## Model architecture: A latent-risk aware machine learning approach for predicting operational success in clinical trials.

Given the diversity of clinical trial outcome targets explored in prior scientific work, and the inherently sequential and interdependent nature of stage-specific success metrics across the clinical drug development lifecycle, this study first establishes a structured definition of clinical development success by explicitly delineating its core metrics, intended to serve as predictive targets within an overall structured modeling framework. Specifically, as shown in Figure 1, success is decomposed into (i) operational success, reflecting the ability to complete the execution of the clinical trial as planned, (ii) scientific success, capturing whether the trial meets its primary endpoint(s) with statistically significant and clinically relevant results, (iii) phase transition, representing advancement of pharmaceutical programs to the next clinical stage, and (iv) regulatory approval, corresponding to successful market authorization.

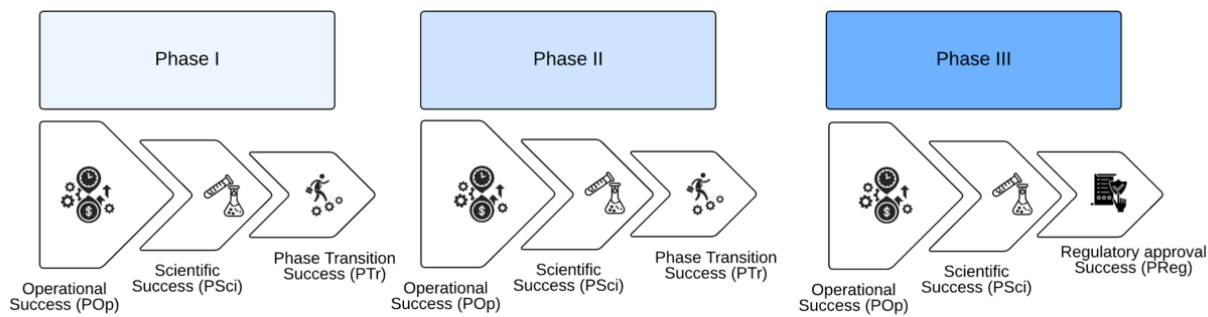

**Figure 1. Stage-specific success metrics and probability-based evaluation across the clinical drug development lifecycle (Phase I–III).** Schematic overview of the sequential and interdependent success metrics considered in this study. Clinical development success is decomposed into four phase-specific predictive probabilities: the probability of operational success (POp), the probability of scientific success (PSci), the probability of phase transition (PTr), and the probability of regulatory approval (PReg), enabling phase-specific modeling and longitudinal risk assessment from early development through downstream milestones.

This study adopts a stepwise modeling strategy that prioritizes the earliest and most proximal success endpoint. The modeling framework therefore first focuses on operational success, which represents the closest observable outcome to trial initiation and constitutes the primary mechanism through which upstream design and execution risks translate into downstream development milestones. Operational success is defined as the ability of a clinical trial to be initiated, conducted, and completed as planned, without major operational failure. In the present study, operational success is operationalized as a binary outcome variable, taking the value (1) when the trial meets all predefined operational criteria and (0) otherwise. This encompasses, but is not limited to:

- Successful study initiation and site activation
- Adequate patient recruitment and retention
- Timely trial execution without premature termination
- Completion of the study through the last patient's last visit milestone and closure of the clinical trial database in accordance with the protocol specifications and operational objectives
- Execution of the trial within planned operational constraints, including resources and cost considerations.

To overcome the limitations inherent to prospective operational modeling, where several high-impact drivers of trial performance are not directly available prior to trial initiation, the proposed framework decomposes operational success into a set of intermediate latent risk drivers. These drivers (such as recruitment performance, dropout risk, protocol amendments, and study duration) are observed retrospectively in historical datasets but remain unknown at the time of trial design. In this framework, they are therefore treated as latent variables that must be predicted from ex-ante information, such as protocol characteristics and investigational drug attributes, and represent intermediate operational mechanisms through which trial design features translate into downstream operational outcomes.

TrialsBank™ includes approximately 20 latent operational risk factors describing different dimensions of trial execution. As a first step toward this broader framework, the present study focuses on a subset of key intermediate drivers selected based on their predictive relevance. Feature-importance analyses indicated that four factors accounted for a substantial proportion of the predictive signal and therefore provided a strong foundation for the initial modeling stage.

Accordingly, as shown in Figure 2, four major latent operational risk factors were modeled as primary determinants of operational success: patient recruitment rate, dropout rate, protocol deviation incidence, and serious adverse event (SAE) occurrence. These factors represent key operational processes affecting trial execution efficiency and data quality yet are not directly measurable prior to trial initiation.

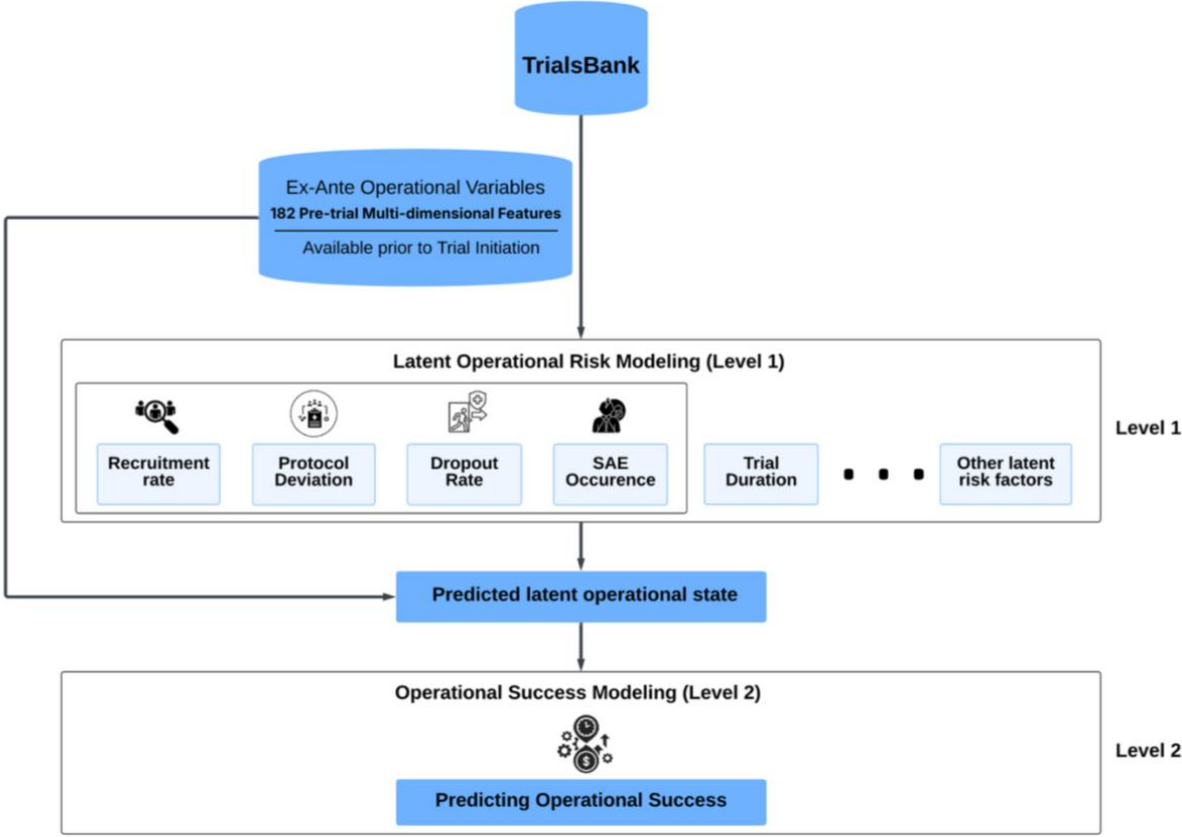

**Figure 2. Hierarchical Latent risk–aware modeling framework for prospective operational success prediction in clinical trials.** Overview of the proposed two-stage modeling pipeline. A shared set of ex-ante drug- and trial-level features available prior to trial initiation is used to predict multiple latent operational risk drivers (Level 1). Among the broader set of operational risk factors considered in the framework, the four boxed factors (recruitment rate, protocol deviation occurrence, dropout rate, and SAE occurrence) were explicitly modeled in the present study. The predicted latent risk outputs are subsequently integrated into a downstream model (Level 2) to estimate the probability of operational success (POp), enabling prospective operational risk forecasting prior to trial initiation.

Each latent risk factor was independently modeled using a dedicated machine learning model trained on a shared set of 182 operational variables available prior to trial initiation. These variables capture trial design characteristics, medical conditions, investigational products, participant populations, and site-level attributes, and other ex-ante operational features. By modeling each latent factor separately, the framework allows each model to learn the predictors specific to that operational risk rather than forcing a single end-to-end model to capture multiple operational risks simultaneously.

The predicted outputs of the four latent risk models were subsequently integrated as inputs into a downstream machine learning model designed to estimate overall operational success of clinical trials. This staged modeling strategy reflects the sequential nature of clinical trial execution; whereby upstream operational processes propagate risk to downstream outcomes. This modular modeling strategy further mitigates overfitting by constraining each model to learn factor-specific patterns rather than optimizing a single end-to-end objective. Indeed, by explicitly modeling these intermediate operational drivers, the approach improves interpretability and aligns predictive outputs with actionable levers available to trial designers prior to study initiation.

## Machine Learning Models

### Data Splitting Strategy

A staged data-splitting strategy, as shown in Figure 3, was implemented to prevent information leakage while enabling robust evaluation of the two-step modeling pipeline. For each clinical phase, the original dataset was first partitioned into three disjoint subsets: a Level 1 training set (40%), a Level 1 validation set (50%), and an independent inference test set (10%), held out from all model development.

The Level 1 validation set was intentionally defined as a larger partition because it serves a dual purpose: (i) evaluating Level 1 model performance, and (ii) generating out-of-sample predictions used as additional input features for Level 2 model training. Specifically, Level 1 models were trained exclusively on the Level 1 training set and then applied to the Level 1 validation set to produce holdout predictions. These predicted latent risk variables were then combined with the original set of ex-ante trial and drug features to form the input space of the Level 2 models. This design ensures that Level 2 learning relies only on out-of-sample Level 1 predictions while preserving the full set of baseline features, thereby approximating real-world deployment conditions and preventing information leakage.

Finally, a fully independent inference test set was reserved to assess the end-to-end performance and generalization of the complete pipeline. This test split was never used for training, model selection, or feature generation, and therefore provides an unbiased estimate of real-world performance under prospective inference settings.

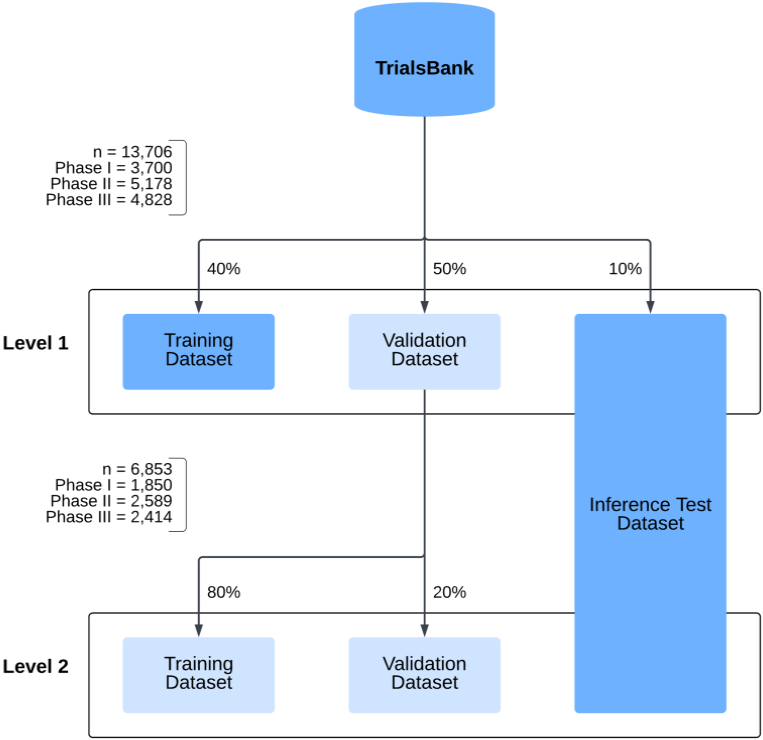

**Figure 3. Staged data-splitting strategy for two-level model training and unbiased inference evaluation.** Schematic of the staged data-splitting protocol used to evaluate the two-step modeling pipeline while preventing information leakage. For each clinical phase, TrialsBank™ was first partitioned into Level 1 training, Level 1 validation, and an independent inference test set. Level 1 models were trained on the training split and applied to the Level 1 validation split to generate out-of-sample predictions, which were subsequently used as input features for Level 2 model training and validation.

The inference test set was held out from all training and model selection steps to provide an unbiased estimate of deployment-like performance.

Missing Data Handling and Imputation

In addition to the construction of a high-quality, curated, context-aware, and AI-ready clinical trial dataset, systematic procedures were implemented to address missing data across selected variables. Missingness mechanisms were first assessed to distinguish between data missing completely at random (MCAR), missing at random (MAR), and missing not at random (MNAR)[34]. Imputation strategies were subsequently selected based on both the identified missingness mechanism and the statistical nature of each variable. Accordingly, standard statistical approaches, including mean and most-frequent-value imputation, were applied where appropriate, while more advanced methods, such as k-nearest neighbors (KNN)–based imputation, were employed for variables exhibiting complex dependency structures.

To assess the robustness of predictive performance to missing data assumptions, all modeling experiments were conducted under two preprocessing conditions: (i) using the selected imputation strategy, and (ii) without imputation, relying only on complete-case information or native missing-value handling when supported by the algorithm.

Model Benchmarking and Training Protocol

Model training was conducted under a controlled benchmarking protocol across Phase I–III datasets. For each prediction task and each phase, three machine learning algorithms were trained and compared: XGBoost, CatBoost, and Explainable Boosting Machines (EBM). These models were selected to jointly capture strong predictive performance, robustness to heterogeneous clinical trial features, and interpretability. Training, hyperparameter tuning, and model selection were applied consistently across algorithms to ensure fair comparison.

Discretization of Continuous Latent Risk Targets

In several applied ML settings, continuous regression targets are discretized into a fixed number of classes, turning the task into a classification problem. This strategy has been reported to improve performance, robustness to noisy targets, and model regularization, while mitigating overly smooth regression outputs[35]. In our context, two of the intermediate targets were additionally reformulated as classification problems through discretization into a fixed number of ordinal categories.

| Latent Risk Factor | Operational definition | Original target format | Final target encoding |
| --- | --- | --- | --- |
| **Protocol Deviations** | Occurrence of at least one protocol deviation during the clinical study. | Binary (True / False) | Binary (True / False) |
| **Serious Adverse Event (SAE) occurrence** | Occurrence of at least one serious adverse event (SAE) during the clinical study. | Binary (True / False) | Binary (True / False) |
| **Recruitment Rate** | Enrollment performance relative to the planned recruitment target for the study. | Continuous (number of enrolled participants) | Ordinal multi-class (deviation from target enrollment)<br>• Above target<br>• On target (±5% deviation)<br>• Below target (5–30% below target)<br>• Severely below target (>30% below target) |
| **Dropout Rate** | Proportion of enrolled participants who discontinued participation prior to study completion | Continuous (number of dropouts, normalized rate: dropouts/enrolled x100) | Ordinal multi-class<br>• No dropout (0%)<br>• Low dropout (<10%)<br>• Moderate dropout (10–40%)<br>• High dropout (>40%) |

**Table 1. Encoding of latent operational risk targets for model training.** This table summarizes the four latent operational risk factors modeled in this study, including their operational definitions, original target formats, and final encoding used for machine learning. Binary targets (protocol deviations and SAE occurrence) were modeled as two-class outcomes, while continuous targets (recruitment performance and dropout rate) were discretized into ordinal multi-class categories to support robust and consistent predictive learning across trials.

Experimental Conditions

Overall, the machine learning experiments were conducted across multiple controlled dimensions: (i) phase-specific training (Phase I, II, III), (ii) algorithmic benchmarking (XGBoost, CatBoost, EBM), (iii) multiple operational targets including latent risk drivers and operational success, and (iv) preprocessing variants with and without imputation. This experimental design enables systematic assessment of model performance, robustness, and interpretability under realistic operational constraints.

# Main Results

## Relationship between latent risk prediction accuracy and operational success prediction

On the phase II inference test set (n=520 clinical trials), we analyzed the composition of trials according to the proportion of the four latent risk factors that were correctly predicted (0%, 25%, 33%, 50%, 67%, 75%, and 100%) and the resulting operational success prediction accuracy. Overall, as shown in Figure 4, higher agreement on latent risk factor predictions was associated with improved operational success prediction performance. In particular, among trials for which all latent risk factors were correctly predicted (100%), operational success was correctly classified in 92% of cases (100/109), with only 8% misclassified (9/109).

Importantly, strong operational success prediction performance remained achievable even when latent risk predictions were imperfect. For example, among trials with 0% correctly predicted latent factors (n=45), operational success was still correctly predicted for 67% of studies (30/45). These findings suggest that while accurate prediction of latent operational drivers substantially strengthens end-to-end performance, the downstream model can partially compensate for upstream uncertainty, supporting robust inference under realistic conditions where intermediate risk factors may be noisy or partially missing.

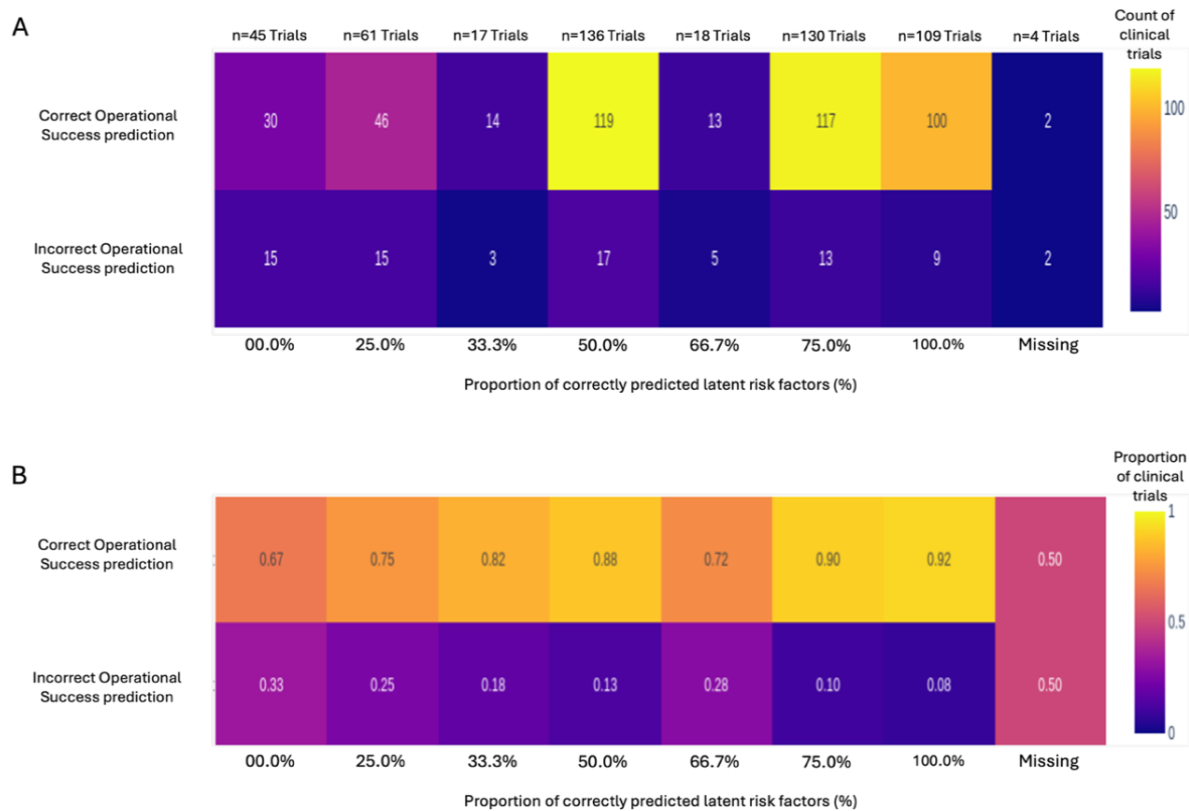

**Figure 4. End-to-end operational success prediction performance as a function of latent risk prediction agreement.**
**(A)** Number of phase II inference test trials grouped by the proportion of correctly predicted latent risk factors (out of four) and whether operational success was correctly predicted. Column headers indicate the total number of trials per agreement level (n). **(B)** Column-normalized proportions showing, for each agreement level, the fraction of trials with correct versus incorrect operational success predictions.

## Sensitivity analysis: impact of each latent risk prediction errors on downstream operational success

To better understand how uncertainty in individual latent operational risk factors propagates to the downstream operational success classifier, we analyzed the Phase II inference test set by stratifying trials according to the prediction error of each latent factor. For ordinal multi-class targets (dropout and recruitment deviation), error was quantified as the absolute distance between the predicted and true class (distance = 0 indicates correct classification, whereas larger distances reflect more severe misclassification). For binary targets (protocol deviation and SAE occurrence), trials were grouped based on correct versus incorrect latent factor prediction.

Overall, as shown in Figure 5 (A, C), the downstream operational success model exhibited clear sensitivity to errors in recruitment- and dropout-related latent factors. In particular, operational success prediction accuracy was highest when dropout was correctly classified (distance = 0; ~90% correct operational success prediction) and decreased when dropout misclassification was severe (distance = 3; ~70%). A similar pattern was observed for recruitment deviation, where correct recruitment class prediction (distance = 0) was associated with ~90% correct operational success prediction, whereas large recruitment misclassification (distance = 3) reduced accuracy to ~62%.

By contrast, as shown in Figure 5 (B, D), protocol deviation showed limited influence on downstream performance, as operational success prediction accuracy remained relatively stable regardless of whether protocol deviation was correctly predicted (~88%). SAE occurrence demonstrated a more moderate effect, with operational success prediction accuracy ranging between ~82% when SAE was correctly predicted and ~87% when misclassified. Finally, trials with missing latent factor ground-truth labels (NA category) consistently exhibited lower operational success prediction accuracy, highlighting the importance of complete and reliable intermediate risk annotations for end-to-end model performance.

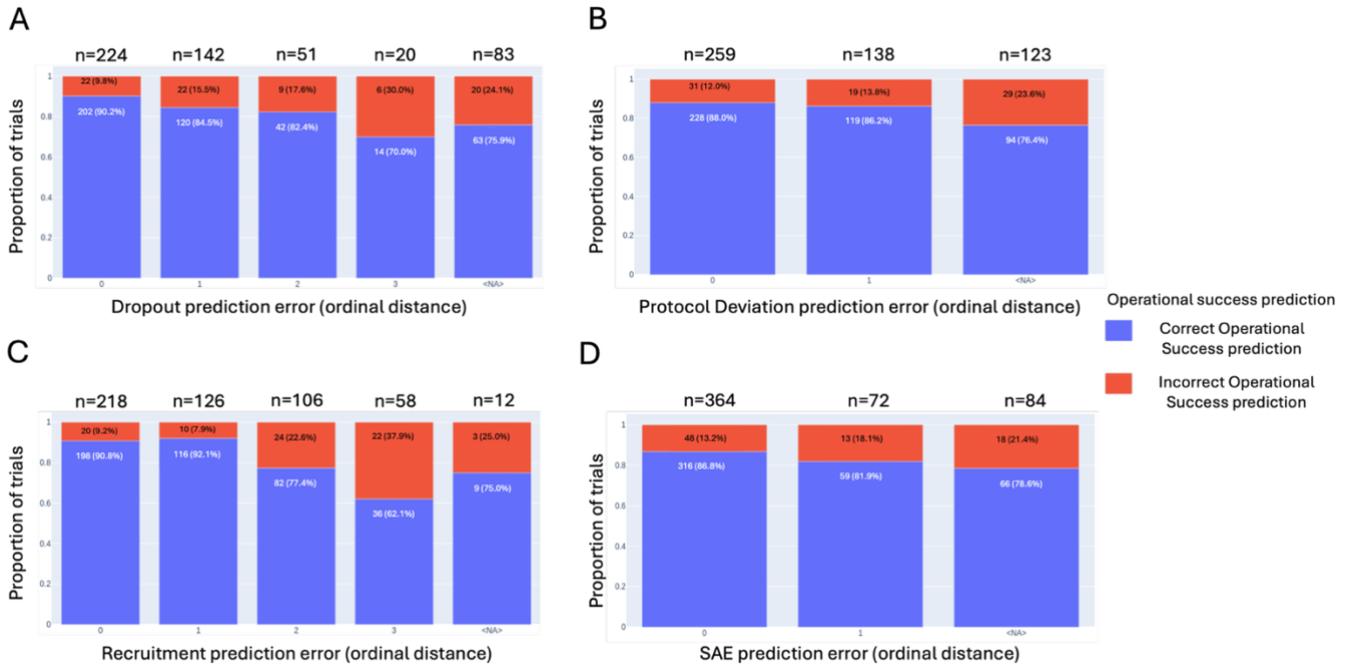

**Figure 5. Relationship between latent risk prediction error and operational success prediction performance (Phase II).**
Stacked bar plots showing, for each latent operational risk factor, the proportion of Phase II inference trials for which operational success was correctly predicted (blue) versus incorrectly predicted (red), stratified by the latent factor prediction error expressed as ordinal distance (0 = correct class, higher values indicate larger misclassification distance). Column headers report the total number of trials in each error category (n). **(A)** Dropout rate prediction error. **(B)** Protocol deviation prediction error. **(C)** Recruitment (accrual deviation) prediction error. **(D)** Serious adverse event (SAE) prediction error. "NA" indicates trials for which the corresponding latent risk target was unavailable.

## Operational success prediction performance across phases and inference settings

To evaluate the predictive performance of the proposed latent risk–aware framework, we conducted a comparative analysis of operational success prediction under multiple experimental settings, specifically on the Level 2 prediction of overall operational trial success within the hierarchical modeling framework.

First, we assessed the impact of incorporating an additional latent operational driver by comparing a baseline configuration using three latent risk factors against an enhanced configuration using four latent risk factors. This comparison was performed on Phase II and Phase III datasets to quantify performance gains attributable to the expanded latent risk representation.

Across Phase II and Phase III, as shown in Table 2, the optimized models substantially outperformed the baseline configuration. In Phase II, the baseline model achieved an accuracy of 0.76, whereas the optimized configurations reached 0.87 accuracy. Notably, the optimized four–latent risk factor model improved Recall (Class 0) from 0.33 to 0.72, while maintaining strong performance for Class 1 with Recall = 0.93. Precision also increased from 0.70 to 0.82 for Class 0, and from 0.77 to 0.89 for Class 1.

In Phase III, the baseline model achieved an accuracy of 0.82 but exhibited low sensitivity for failures (Recall Class 0 = 0.14). Both optimized configurations improved overall accuracy to 0.89, with the four–latent risk factor model increasing Recall (Class 0) to 0.60, while preserving high Recall (Class 1 = 0.96) and strong precision (Precision Class 1 = 0.90).

To evaluate generalization under deployment-like conditions, the best-performing configuration (optimized model with four latent risk factors) was evaluated on an independent inference test set. The model maintained strong predictive performance, achieving an accuracy of 0.84 (Phase II inference) and 0.89 (Phase III inference). In particular, the inference results remained stable for the success class (Recall Class 1 = 0.91 in Phase II inference and 0.96 in Phase III inference), while maintaining moderate sensitivity for the failure class (Recall Class 0 = 0.69 and 0.54, respectively). Overall, these findings indicate that incorporating four latent operational risk factors improves discrimination between successful and unsuccessful trials, while maintaining robust generalization from experimentation to inference settings.

| Dataset | Modeling configuration | Accuracy | Recall (Class 0*) | Recall (Class 1**) | Precision (Class 0) | Precision (Class 1) |
|---|---|---|---|---|---|---|
| **Phase II (Experimentation)** | Baseline (3 latent risk factors) | 0.76 | 0.33 | 0.94 | 0.70 | 0.77 |
| | Optimized model (3 latent risk factors) | 0.87 | 0.71 | **0.94** | **0.84** | 0.88 |
| | **Optimized model (4 latent risk factors)** | **0.87** | **0.72** | 0.93 | 0.82 | **0.89** |
| **Phase III (Experimentation)** | Baseline (3 latent risk factors) | 0.82 | 0.14 | 0.99 | 0.88 | 0.81 |
| | Optimized model (3 latent risk factors) | 0.89 | 0.59 | **0.97** | **0.82** | 0.90 |
| | **Optimized model (4 latent risk factors)** | **0.89** | **0.60** | 0.96 | 0.81 | **0.90** |
| **Inference test set (phase II)** | **Optimized model (4 latent risk factors)** | **0.84** | **0.69** | **0.91** | **0.73** | **0.89** |
| **Inference test set (phase III)** | **Optimized model (4 latent risk factors)** | **0.89** | **0.54** | **0.96** | **0.75** | **0.92** |

**Table 2. reports Phase II–III comparative results between the three-factor baseline and the optimized latent risk model for predicting overall operational trial success at Level 2 of the hierarchical modeling framework, including inference performance.**

\* Class 0 = Operational success not achieved
\*\* Class 1 = Operational success achieved

Finally, to assess reproducibility and phase-wise consistency of the proposed framework, the operational success prediction task was replicated across Phase I–III using the same modeling pipeline. As shown in Table 3, the downstream Level 2 EBM model (best performing ML model) achieved consistently strong performance across phases, with F1-scores of 0.93, 0.92, and 0.91 in Phase I, Phase II, and Phase III, respectively. Similarly, classification accuracy remained stable across phases (0.89, 0.85, and 0.87, respectively), indicating that the proposed approach maintains robust predictive performance throughout the clinical development lifecycle.

| Model | Phase I (F1-score) | Phase I (Accuracy) | Phase II (F1-score) | Phase II (Accuracy) | Phase III (F1-score) | Phase III (Accuracy) |
|---|---|---|---|---|---|---|
| EBM (Level 2 operational success) | 0.93 | 0.89 | 0.92 | 0.85 | 0.91 | 0.87 |

**Table 3. reports final model performance for Level 2 operational success prediction across clinical phases.**

# Discussion

This study proposes and evaluates a latent risk–aware machine learning framework for forecasting clinical trial operational success using TrialsBank™, covering Phase I–III studies across all therapeutic areas. Overall, the results demonstrate that multi-dimensional features spanning investigational therapy characteristics, disease and indication attributes, trial design and protocol characteristics, outcome and endpoint specifications, study participant characteristics, sponsor characteristics, and site-level characteristics available prior to trial initiation can serve as highly informative predictors of downstream operational success, supporting the feasibility of data-driven operational risk assessment early in the clinical development lifecycle. Importantly, predictive performance remained consistent across study phases, suggesting that the proposed modeling pipeline captures phase-agnostic patterns that generalize from Phase I to Phase III settings.

A key contribution of this work lies in the stage-specific modeling strategy, where operational success prediction is strengthened by integrating upstream models of intermediate operational variables that are typically unobserved ex-ante yet significantly influence overall trial success (e.g., recruitment performance, dropout rate, protocol deviation occurrence, and SAE incidence). The strong downstream performance achieved through this integration highlights the value of decomposing clinical trial success into sequential success metrics rather than relying solely on a single end-to-end outcome. Importantly, this design enables prospective forecasting by generating intermediate operational predictions before trial completion, allowing stakeholders to anticipate emerging risks and proactively implement mitigation strategies. By explicitly modeling these intermediate drivers, the framework supports practical adoption in real-world clinical development workflows where operational variables must be anticipated, monitored, and managed ahead of time.

Looking forward, the results motivate multiple extensions. The stage-specific framework can be expanded to incorporate the ~20 additional operational variables available in TrialsBank™ (e.g., trial duration), enabling more granular characterization of operational risk and strengthening the predictive signal available to downstream success models.

Moreover, from a modeling standpoint, the strong performance achieved for operational success prediction represents an initial step toward a broader pipeline capable of forecasting additional sequential outcomes, including scientific success, phase transition, and regulatory approval. This extension is particularly enabled by TrialsBank™, which longitudinally links clinical studies across Phase I through regulatory decision-making, thereby supporting longitudinal analysis of complete development trajectories and allowing risk to be quantified at each stage-specific success metric at the program level.

In subsequent work, an important extension of this predictive framework will be to translate operational risk forecasts into actionable trial optimization recommendations, with the objective of identifying protocol and design elements that maximize operational success while accounting for real-world constraints such as timelines, feasibility, regulatory requirements and budget. Achieving this goal will likely require integrating causal modeling to distinguish intervention-relevant drivers from purely correlational associations and to support decision-making under realistic implementation constraints.

From an industry perspective, the proposed framework may support multiple decision layers across the drug clinical development lifecycle. For sponsors and biotechnology companies, prospective operational risk forecasting can inform early feasibility assessments, country and site strategy design, scenario-based protocol optimization, and portfolio-level prioritization across competing trial concepts. By quantifying the probability of operational success prior to trial initiation, sponsors may better align timeline assumptions, budget projections, and risk mitigation plans with realistic execution conditions. For CROs, such predictive insights may serve as structured inputs into feasibility validation, delivery planning, and Risk-Based Quality Management (RBQM) design. Predicted latent risk drivers, such as recruitment underperformance, elevated dropout risk, or protocol deviation likelihood, can inform monitoring strategy, centralized oversight focus, resource allocation, and targeted mitigation planning. Importantly, these tools are best positioned as decision-support inputs within established governance frameworks, complementing rather than replacing expert judgment and quality management systems.

When integrated within documented oversight processes and critical-to-quality risk assessments, such predictive frameworks may enhance transparency, proportionality, and proactive risk mitigation in alignment with contemporary regulatory expectations for sponsor oversight.

More broadly, these findings highlight the value of TrialsBank™ as an expert-curated, AI-ready data infrastructure for clinical development analytics. By integrating scientific, pharmacologic, operational, regulatory, and economic information across longitudinal drug development trajectories, the database integrates trial-, protocol-, sponsor-, and therapy-level data and reconstructs clinical development pathways from Phase I through regulatory decision-making. As such, it provides a scalable foundation for next-generation clinical development intelligence systems, enabling predictive, causal, and optimization frameworks aimed at supporting data-driven decision-making throughout the clinical development lifecycle.

## Conclusion

This work demonstrates that operational success in clinical trials can be forecasted with strong predictive performance using a latent risk–aware modeling pipeline trained on TrialsBank™ Phase I–III data. By combining ex-ante trial and drug characteristics with intermediate operational risk predictions, the proposed approach achieved consistent results across phases and maintained robust performance under independent inference evaluation. Overall, these findings support the feasibility of prospective, data-driven operational risk forecasting as a scalable foundation for clinical development decision-support applications.

## Acknowledgments

This work was primarily funded by Groupe Sorintellis Inc., with additional support from Mitacs through projects IT33089, IT37571, and partial funding from IVADO (Institut de Valorisation des Données) and Biotalent Canada. The authors thank all domain experts and collaborators who contributed to the construction and validation of TrialsBank™ through data curation, expert annotation, and methodological discussions. We also acknowledge the valuable support of industrial partners from Pfizer Global, Vantage BioTrials, CDS monitoring, for their insights into operational feasibility and real-world clinical trial practices. The authors further acknowledge the contributions of graduate interns and trainees whose efforts supported data preparation and database development.

## Declaration of interests

E.P. is employee and shareholders of Groupe Sorintellis Inc.
I.H. and J.H. are employees of Groupe Sorintellis Inc.
K.L. is a consultant for Groupe Sorintellis Inc.
S.E. and V.B. are affiliated with Vantage BioTrials, an industrial partner of Groupe Sorintellis Inc.
All other authors declare no competing financial or non-financial interests.

E.P. reports pending patent applications related to the methods described in this study (2023/WO2023245301A1).
E.P. and I.H. report pending patent applications related to the methods described in this study (2024/WO2025129360A1).